\pdfoutput=1

\documentclass[11pt]{article}

\usepackage{ACL2023}

\usepackage{times}
\usepackage{latexsym}

\usepackage{booktabs}
\usepackage{amsmath}
\usepackage{graphicx}
\usepackage[normalem]{ulem}
\useunder{\uline}{\ul}{}
\usepackage{float}
\usepackage{multirow}
\usepackage{bbold}
\usepackage{hhline}
\usepackage{xcolor}
\usepackage{cleveref}
\usepackage{adjustbox}
\definecolor{mygreen}{RGB}{76, 153, 0}

\DeclareMathOperator*{\cossim}{sim}
\DeclareMathOperator*{\avg}{avg}

\usepackage{todonotes}

\usepackage[T1]{fontenc}

\usepackage[utf8]{inputenc}

\usepackage{microtype}

\usepackage{inconsolata}

\title{Arithmetic-Based Pretraining -- Improving Numeracy of Pretrained Language Models}

\author{Dominic Petrak$^{\dag}$ , Nafise Sadat Moosavi$^{\ddag}$, Iryna Gurevych$^{\dag}$\\
  \\
  $^{\dag}$Ubiquitous Knowledge Processing Lab (UKP Lab), \\
  Department of Computer Science and Hessian Center for AI (hessian.AI), \\ 
  Technical University of Darmstadt, Germany\\
  \url{https://www.ukp.tu-darmstadt.de}\\
  $^{\ddag}$Department of Computer Science, The University of Sheffield, UK\\
  }

\begin{document}
\raggedbottom
\maketitle
\begin{abstract}State-of-the-art pretrained language models tend to perform below their capabilities when applied out-of-the-box on tasks that require understanding and working with numbers. Recent work suggests two main reasons for this: (1) popular tokenisation algorithms have limited expressiveness for numbers, and (2) common pretraining objectives do not target numeracy. Approaches that address these shortcomings usually require architectural changes or pretraining from scratch. In this paper, we propose a new extended pretraining approach called Arithmetic-Based Pretraining that jointly addresses both in one extended pretraining step without requiring architectural changes or pretraining from scratch. Arithmetic-Based Pretraining combines contrastive learning to improve the number representation, and a novel extended pretraining objective called Inferable Number Prediction Task to improve numeracy. Our experiments show the effectiveness of Arithmetic-Based Pretraining in three different tasks that require improved numeracy, i.e., reading comprehension in the DROP dataset, inference-on-tables in the InfoTabs dataset, and table-to-text generation in the WikiBio and SciGen datasets\footnote{Code, data, and models trained using Arithmetic-Based Pretraining are available here: \url{https://github.com/UKPLab/starsem2023-arithmetic-based-pretraining}.}.

\end{abstract}

\section{Introduction}

Numbers are ubiquitous in natural language. Therefore, understanding and working with numbers (usually referred to as numeracy) is a critical capability for pretrained language models such as BART~\cite{lewis2019bart} or T5~\cite{raffel_t5}, cornerstones of modern NLP, in order to utilize quantitative information for various NLP tasks. Recent works question whether these models meet this requirement out-of-the-box~\cite{wallace2019nlp, zhang2020language}: Common pretraining objectives such as the denoising autoencoder of BART~\cite{lewis2019bart}, the masked language modeling objective of BERT~\cite{devlin2018bert}, or the span-corruption objective of T5~\cite{raffel_t5}, are designed for understanding structure and semantic meaning of language and not to learn working with numbers. Furthermore, commonly used subword-based tokenisation algorithms such as Byte Pair Encoding \cite{sennrich2015neural} or WordPiece~\cite{wu2016google} are designed to handle patterns that are frequently observed during training, which is disadvantageous for numbers. For instance, 0.72 and 0.73 are two similar numbers. They should be processed similarly, but according to their frequency in the pretraining data they might be tokenised very differently, e.g., [0, ., 72] and [0, ., 7, 3], which will have an impact on their representation in embedding space. To address these shortcomings, various approaches have been proposed recently. However, most of them introduce additional components or rely on predefined features that limit their application, e.g., they are only applicable in a specific task like reading comprehension~\cite{andor-etal-2019-giving, geva2020injecting} or require architectural changes~\cite{herzig-etal-2020-tapas}.

In this paper, we propose a new extended pretraining approach called Arithmetic-Based Pretraining that targets both shortcomings for pretrained language models in one extended pretraining step without introducing new components or requiring pretraining from scratch. 
It consists of:

\begin{itemize}
    \item A contrastive loss that combines subword-based with character-level tokenisation to improve the representation of numbers.
    \item A denoising pretraining objective, called the Inferable Number Prediction Task, to improve the model's capability of working with \mbox{numbers}.
\end{itemize}

Our experiments show that Arithmetic-Based Pretraining has a positive impact on BART~\cite{lewis2019bart}, T5~\cite{raffel_t5} and Flan-T5~\cite{chung_flan_t5} in various tasks. It improves the accuracy in case of reading comprehension and inference-on-tables, and the factual correctness in case of table-to-text generation.


\section{Related Work}

\paragraph{Number Representations in Language Models.}
\label{sub_sec:num_reps}
State-of-the-art language models like BART~\cite{lewis2019bart} or T5~\cite{raffel_t5} use subword-based tokenisation algorithms (such as Byte Pair Encoding~\cite{sennrich2015neural}) to build vocabularies based on frequently observed sequences in a text corpus. While this is effective for common words, it is problematic for numbers. 
In an extensive study, \newcite{wallace2019nlp} shows that models using character-level tokenisation, such as ELMo~\cite{Peters:2018}, usually achieve better results in numerical probing tasks and extrapolate better to unseen numbers compared to models using subword-based tokenisation. \newcite{thawani2021representing}, \newcite{peng2021mathbert} and \newcite{zhang2020language} report similar findings. In our work, we use the character-level tokenisation for numbers to address this shortcoming in BART, T5, and Flan-T5~\cite{chung_flan_t5}.

\paragraph{Approaches for Improving Numeracy.}
Numeracy requires to understand and work with numbers, i.e., to do artihmetic operations, in order to generate the expected result. 
To improve this capability, recent approaches propose pretraining from scratch or architectural changes to tailor pretrained language models towards specific tasks. TAPAS~\cite{herzig-etal-2020-tapas} targets question answering with tabular data. It is pretrained from scratch and extends BERT~\cite{devlin2018bert} by introducing additional embeddings for capturing tabular structure. 
GenBERT~\cite{geva2020injecting} reuses a pretrained BERT model and adds a decoder on top. It is then further trained using math word problems and arithmetic operations for (1) incorporating the character-level tokenisation for numbers, and (2) to improve the numerical reasoning skills. It achieves state-of-the-art results in the DROP~\cite{dua-etal-2019-drop} and SQUAD~\cite{rajpurkar-etal-2016-squad} datasets. 
\citet{andor-etal-2019-giving} also reuses the pretrained BERT model and targets reading comprehension. They add a new layer on top that predicts and executes arithmetic operations.
\citet{suadaatowards} target table-to-text generation and propose a framework that uses the template-guided text generation from \newcite{kale-rastogi-2020-template} to inject pre-executed numerical operations into the pretrained GPT-2~\cite{radford2019language} and T5~\cite{raffel_t5} models.

In their experiments, all of these works show that much of their performance improvements are due to specific design decisions or multi-level pretraining setups which result in new or task-specific models. With Arithmetic-Based Pretraining, we propose an approach that improves a model's numeracy with just one extended pretraining step and without changing its architecture.

\paragraph{Domain-Adaptive Pretraining.}
The idea of domain-adaptive pretraining is to bridge the gap between the vocabulary of a model's original pretraining corpus and the target domain by continuing pretraining using in-domain data~\cite{gururangan2020don}. 
In this work, we propose the Inferable Number Prediction Task which is similar to domain-adaptive pretraining if the data used is from the same domain as that of finetuning. However, we show that this is not the only reason for performance improvements (Section~\ref{sec:ood_data}).

\paragraph{Contrastive Learning.}
Contrastive learning is a general way to learn to map vector representations of similar data points (usually called \textit{anchor} and \textit{positive}) close to each other while pushing non-similar data points apart. In NLP, it is commonly used for learning sentence representations~\cite{kim-etal-2021-self, giorgi-etal-2021-declutr} or semantic similarities~\cite{wang-etal-2021-cline}. In this work, we use contrastive learning to improve the representation of numbers.

\section{Arithmetic-Based Pretraining}
In this section, we propose Arithmetic-Based Pretraining. It combines different tokenisation algorithms, i.e., character-level and subword-based, with contrastive learning to improve the representation of numbers in pretrained language models (Section~\ref{sec:contrastive}), while training on the Inferable Number Prediction Task (Section~\ref{sec:mnp}) to improve the capability of working with numbers. Section~\ref{sect:loss} describes the joint loss function.


\subsection{Contrastive Learning} \label{sec:contrastive}
We propose to use a contrastive loss as additional training signal to improve the representation of numbers.
For example, the model should learn a similar representation for the number 108.89, whether it is initially tokenised as [1, 0, 8, ., 8, 9] (character-level) or [10, 8, ., 89] (subword-based). If a number frequently occurs in the pretraining corpus, its corresponding subword-based encoding may be more informative. If this is not the case, its character-level tokenisation may be more informative. Therefore, our motivation is to benefit from both embedding spaces for learning better number representations. 
For implementation, we use the Multiple Negative Ranking Loss as proposed by \newcite{henderson2017efficient}\footnote{We use the implementation from the sentence-transformer library \citep{reimers-gurevych-2019-sentence}.}:

\begin{equation} 
\label{eq:mnrl}  
    \mathcal{L}_{C} =- \frac{1}{N}\sum_{i=1}^{N}\frac{e^{\cossim(\avg(\hat{p}_i), \avg(\hat{p\prime}_i))}}{\sum_j e^{\cossim(\avg(\hat{p}_i), \avg(\hat{p}_{neg}))}}  
\end{equation}

For the contrastive loss, we consider all numbers in the batch independently of the input sequences. Each number is used twice, once in character-level tokenisation (anchor), and once in subword-based tokenisation\footnote{Note that we use both only for Arithmetic-Based Pretraining. For finetuning and during inference, we only use character-level tokenisation for numbers.}. Assume $p$ is a list of all numbers in the batch in character-level tokenisation. $p\prime$ is a list of all numbers in the batch in subword-based tokenisation. We consider $p_i$ and $p\prime_i$ as a positive pair. Every other number in $p$ and $p\prime$ is considered as negative sample to $p_i$ (denoted as $p_{neg}$). $\hat{p_i}$, $\hat{p\prime}_{i}$, and $\hat{p}_{neg}$ are the corresponding embeddings after the encoder pass. $sim$ represents the cosine similarity and $avg$ represents the mean-average of the embedding. Averaging is a simple and effective form of aggregation which is necessary at this point, as the numbers are split into multiple tokens during tokenisation. 

\subsection{The Inferable Number Prediction Task} \label{sec:mnp}
The Inferable Number Prediction Task is a variation of the classic masked language modeling objective~\cite{devlin2018bert}, but aims on improving a model's capability on working with numbers by focusing on data that requires arithmetic operations. The task consists of input $C$ and the corresponding target sequence $D$. $C$ consists of a pair of text sequences, $C_1$ and $C_2$, that are separated with a special character. $C_2$ equals to $D$, but contains a masked number that can be inferred from $C_1$.
Given $C$, the task is to reconstruct $D$ by correctly predicting the masked number in $C_2$\footnote{Preliminary experiments revealed that just reconstructing the masked number, without its context, has a negative impact on a model's text generation capabilities.}. For instance, for the task of table-to-text generation, $C$ consists of the linearized form of the input table ($C_1$) and its description with one masked number ($C_2$).
We select data with the following criteria:
\begin{itemize}
\item $D$ ($C_2$ in $C$) and $C_1$ should have at least one overlapping entity, e.g., $D$ should contain at least one of the entities that appear in the row or column headers of $C_1$ if $C_1$ is a table. This ensures that $D$ is relevant to the information given in $C_1$. 
\item $D$ ($C_2$ in $C$) should contain at least one number that either occurs in $C_1$ or is inferable by summation, subtraction, multiplication, division or \mbox{ordering}. This ensures that the masked number in $C_2$ is arithmetically related to the numbers given in $C_1$.
\end{itemize}  
Next, we reduce $C$ to the necessary information. If $C_1$ is an extensive text or paragraph, we apply each of these heuristics to each of the sentences and retain only the matching ones (the same applies to $C_2$). If $C_1$ is a table, we remove rows and columns that do not share entities with $C_2$ (see Appendix~\ref{sec:masked_task_examples} for further details and illustrations). 

For training, we use the cross-entropy loss function:

\begin{equation} \label{eq:cross_entropy} 
    \mathcal{L}_{INP}(x, y) = \frac{1}{N} \sum_{n=1}^{N} -\log \left (\frac{e^{(x_n,_{y_n})}}{\sum_{k=1}^{K}e^{(x_{n,k})}} \right )    
\end{equation} 
where $x$ represents the logits of the predicted input sequence, and $y=y_1,...,y_N$ represents the indices of the tokens of the output sequence. $N$ is the size of the target sequence. $x_n,_{y_n}$ is the logit of the $x_n$ token corresponding to the output token $y_n$. $K$ is the size of the model's vocabulary.

\subsection{Joint Loss Function}
\label{sect:loss}
We combine the contrastive loss $\mathcal{L}_{C}$ (Equation~\ref{eq:mnrl}) and the loss for the Inferable Number Prediction Task $\mathcal{L}_{INP}$ (Equation~\ref{eq:cross_entropy}) as weighted sum in a joint loss function:
\begin{equation} \label{eq:masked_overall_loss}  
    \mathcal{L} = \frac{\mathcal{L}_{C}}{2} + \frac{\mathcal{L}_{INP}}{2}  
\end{equation}
\section{Experimental Setup}

We implement our approach using Python 3.10, PyTorch~\cite{paszkepytorch} and Huggingface~\cite{wolf-etal-2020-transformers}. 
As pretrained language models, we use the large variant of BART~\cite{lewis2019bart} and the base variant of T5~\cite{raffel_t5} and Flan-T5~\cite{chung_flan_t5} as provided by the Huggingface platform (see Appendix~\ref{sec:appendix_3} for details on hyperparameters)\footnote{We could not use the large variant of T5 and Flan-T5 due to hardware limitations (each model has 770M parameters).}. All models are pretrained Transformer-based encoder-decoder models, but different in size. BART-large consists of a total of $24$ layers and 406M parameters. T5-base and Flan-T5-base consist of $12$ layers and 220M parameters. Flan-T5 is based on T5, but trained on more tasks, e.g., arithmetic reasoning, and chain-of-thought data (instructions). It significantly improves the results of the original model in many tasks~\cite{chung_flan_t5}. We conduct all experiments on a Tesla V100-SXM3 GPU with $32$ GB memory. For experiments using table-to-text datasets, we represent tables as linearized sequence. We report the results of the best single runs.

\subsection{Original Datasets}
\label{sect:dataset}

\paragraph{Reading Comprehension.}
The task of reading comprehension is to answer a question by reasoning over a related text passage. DROP~\cite{dua-etal-2019-drop} is such a dataset. It contains over 96,567 open-domain question-answer pairs and 6,735 paragraphs. According to the authors, $59.1\%$ of answers consist of numbers and therefore implicitly require performing arithmetic operations to be predicted correctly. Each paragraph consists of $9.19\%$ numbers on average. We split the dev data into two equally-sized subsets and use one for testing. Each subset contains 4,828 question-answer pairs.

\paragraph{Inference-on-Tables.}
Given a premise and a hypothesis, natural language inference (NLI) is the task of deciding whether the hypothesis is entailed, contradictory, or neutral to the premise.
InfoTabs~\cite{gupta-etal-2020-infotabs} extends NLI to using semi-structured data, i.e.,  tables, as hypothesis. It consists of 23,738 hypothesis for 2,540 Wikipedia infoboxes from a variety of domains and provides three different test sets: in-domain, cross-domain, and an adversarial test set. The cross-domain test set uses premises from domains not used for training. The adversarial test set uses a different set of source tables. Furthermore, the wording of hypotheses was slightly changed by expert annotators. According to the authors, InfoTabs requires numerical and temporal reasoning (which implicitly requires performing arithmetic operations) across multiple rows and to a large extent. Each table consists on average of $13,89\%$ numbers.

\paragraph{Table-to-Text Generation.}
Table-to-text generation is the task of summarizing tabular data (which is often numerical) in a descriptive text. It requires to implicitly perform arithmetic operations such as ordering, summation or subtraction, or to capture magnitudes. SciGen~\cite{moosavi2021learning} is a table-to-text generation dataset that requires to generate descriptions for scientific tables\footnote{NumericNLG~\cite{suadaatowards} is a similar dataset. As SciGen~\cite{moosavi2021learning} provides more unsupervised training pairs that we can use for Arithmetic-Based Pretraining, we use SciGen in our experiments.}. It is designed for arithmetic reasoning and consists of 53,136 table-description pairs. Each table consists of $41.55\%$ numbers on average.

WikiBio~\cite{Lebret_EMNLP2016} is a dataset from the biographical domain. It consists of 728,321 table-description pairs. The task is to reproduce the first paragraph of biographical Wikipedia articles, given the corresponding infobox. According to the authors, dates, ages, and other quantities play an important role. Each table consists of $16.83\%$ numbers on average. However, most values can be directly copied from the tables and do not require arithmetic operations.

\subsection{Preprocessing for the Inferable Number Prediction Task}
\label{sec:inp_datasets}
To fulfill the requirements of the Inferable Number Prediction Task, we apply the criterias described in Section~\ref{sec:mnp} to all datasets in an offline preprocessing step. In case of InfoTabs~\cite{gupta-etal-2020-infotabs}, we only use the data labeled with entailed in order to exclude contradictions (see Appendix~\ref{sec:masked_task_examples} for examples and illustrations). Table~\ref{tab:mnp_statistics} shows the resulting datasets.

\begin{table}[!htb]
  \centering
  \resizebox*{0.7\linewidth}{!}{  

    \begin{tabular}{lrrr}
        \multicolumn{1}{c}{\textbf{}} & \multicolumn{1}{c}{\textbf{Train}} & \multicolumn{1}{c}{\textbf{Dev}} & \multicolumn{1}{c}{\textbf{Test}} \\ \hline
        \textbf{SciGen} & 4,859 & 1,473 & 55 \\ 
        \textbf{WikiBio} & 412,053 & 51,424 & 51,657 \\ 
        \textbf{DROP} & 8,336 & 849 & 850 \\ 
        \textbf{InfoTabs} & 1,981 & 1,800 & 1,800 \\ 
        \end{tabular}
}

\caption{Data distribution for the Inferable Number Prediction Task after applying the criterias to the original dataset splits.}
\label{tab:mnp_statistics}
\end{table}

We also find that the resulting datasets have slightly different number-to-word ratios. In the case of DROP~\cite{dua-etal-2019-drop} and InfoTabs, preprocessing increases the portion of numbers up to $18.98\%$ and $17.25\%$ in paragraphs and tables. In the case of WikiBio~\cite{Lebret_EMNLP2016} the ratio remains unchanged and in the case of SciGen~\cite{moosavi2021learning} it reduces the numbers per table to $33.88\%$. 

\begin{table}[!htb]
\centering
\resizebox*{1.0\linewidth}{!}{
\begin{tabular}{lllllll}
 & \textbf{OCC} & \textbf{ORD} & \textbf{SUM} & \textbf{SUB} & \textbf{MUL} & \textbf{DIV} \\ \hline
\textbf{DROP} & \multicolumn{1}{r}{$0.41$} & \multicolumn{1}{r}{$0.32$} & \multicolumn{1}{r}{$0.04$} & \multicolumn{1}{r}{$0.07$} & \multicolumn{1}{r}{$0.13$} & \multicolumn{1}{r}{$0.02$} \\
\textbf{InfoTabs} & \multicolumn{1}{r}{$0.23$} & \multicolumn{1}{r}{$0.34$} & \multicolumn{1}{r}{$0.05$} & \multicolumn{1}{r}{$0.17$} & \multicolumn{1}{r}{$0.15$} & \multicolumn{1}{r}{$0.06$} \\
\textbf{SciGen} & \multicolumn{1}{r}{$0.11$} & \multicolumn{1}{r}{$0.06$} & \multicolumn{1}{r}{$0.03$} & \multicolumn{1}{r}{$0.12$} & \multicolumn{1}{r}{$0.41$} & \multicolumn{1}{r}{$0.27$} \\
\textbf{WikiBio} & \multicolumn{1}{r}{$0.24$} & \multicolumn{1}{r}{$0.38$} & \multicolumn{1}{r}{$0.03$} & \multicolumn{1}{r}{$0.10$} & \multicolumn{1}{r}{$0.20$} & \multicolumn{1}{r}{$0.03$}
\end{tabular}
}
\caption{Distribution of arithmetic operations in the preprocessed datasets.}
\label{tab:arithmetic_operations}
\end{table}

Table~\ref{tab:arithmetic_operations} shows the ratio of samples per dataset that we have identified as being inferable by arithmetic operiations, i.e., occurence (OCC), ordering (ORD), summation (SUM), subtraction (SUB), multiplication (MUL) or division (DIV). Appendix~\ref{sec:masked_task_arithmetic_details} provides a detailed analysis.
\section{Evaluation}
\label{sec:evaluation}
In this section, we evaluate the impact of Arithmetic-Based Pretraining on downstream applications with BART~\cite{lewis2019bart}, T5~\cite{raffel_t5} and Flan-T5~\cite{chung_flan_t5} using in-domain data (Section~\ref{sec:downstream}), and out-of-domain data (Section~\ref{sec:ood_data}). For Arithmetic-Based Pretraining, we use  the preprocessed subsets of the original datasets as described in Section~\ref{sec:inp_datasets}.

\subsection{Evaluation Metrics}
\label{sec:metrics}
For inference-on-tables, we evaluate the results using Exact Match (EM score). For reading comprehension, we additionally use F1 score. The EM score evaluates the prediction accuracy, i.e., if the prediction exactly matches the target. It is the preferred metric for these tasks~\cite{dua-etal-2019-drop, gupta-etal-2020-infotabs}. The F1 score reports the overlap between the prediction and the target. This results in partial reward in cases where the prediction is partially correct.
In case of table-to-text generation, we conduct a human evaluation. This is due to the shortcomings of common automatic metrics for this task, as they are hardly able to assess the correctness of information not directly contained in the source data, i.e., information obtained by reasoning~\cite{moosavi2021learning,chen2020logical,suadaatowards}. We provide the results of the automatic metrics in Appendix~\ref{sec:evaluation_automatic_metrics}.

For all experiments, \emph{Baseline} represents the BART~\cite{lewis2019bart}, T5~\cite{raffel_t5}, and Flan-T5~\cite{chung_flan_t5} model directly finetuned on the corresponding dataset without Arithmetic-Based Pretraining. \emph{Ours} represents these models with Arithmetic-Based Pretraining. We highlight statistically significant improvements of Ours over the respective baseline in the tables (independent two-sample t-test, $p \leq 0.05$).

\subsection{In-Domain Pretraining}
\label{sec:downstream}
This section discusses the results on downstream tasks when using models that are pretrained using Arithmetic-Based Pretraining with in-domain data. For comparison, we will also report the results of the specialised state-of-the-art model for each task. 

\paragraph{Reading Comprehension.}
Table~\ref{tab:final_evaluation_drop} shows the results achieved on DROP~\cite{dua-etal-2019-drop}.

\begin{table}[ht]
\centering
\resizebox*{0.6\linewidth}{!}{  
\begin{tabular}{llrr}
 &  & \multicolumn{1}{c}{\textbf{EM}} & \multicolumn{1}{c}{\textbf{F1}} \\ \hline
\multirow{2}{*}{\textbf{BART}} & Baseline & 36.00 & 39.26 \\ 
 & Ours & \textbf{45.60} & \textbf{49.50} \\ \cline{1-4} 
\multirow{2}{*}{\textbf{T5}} & Baseline & 10.40 & 14.60 \\ 
 & Ours & 11.00 & 15.20 \\ \cline{1-4} 
\multirow{2}{*}{\textbf{Flan-T5}} & Baseline & 46.34 & 94.41 \\ 
 & Ours & \textbf{72.18} & \textbf{97.65} \\ \hhline{====}
 \textbf{QDCAT} & & 85.46 & 88.38 \\
\end{tabular}
}
\caption{Evaluation on the DROP dataset. Our approach outperforms the baseline in all cases.}
\label{tab:final_evaluation_drop}
\end{table}

In all cases, Arithmetic-Based Pretraining improves the results over the baseline. Based on our analysis of the test results, i.e., by comparing the predictions of Baseline with Ours, we find that our approach reduces the incorrectly predicted numbers by $14.27\%$ in case of BART~\cite{lewis2019bart}, $16.62\%$ in case of T5~\cite{raffel_t5}, and $30.56\%$ in case of Flan-T5~\cite{chung_flan_t5}. The results achieved with Flan-T5 even outperform the results reported by \newcite{geva2020injecting} for GenBERT (EM 68.6)\footnote{We also did preliminary experiments with the math word problems dataset provided by Geva et al.~\cite{geva2020injecting} as a first pretraining task but found that this does not improve the results (see Appendix~\ref{sec:preliminary_math_experiments}).}. Regarding the performance differences between BART and T5, we attribute this to the difference in model size. In this context, the performance difference between BART and Flan-T5 is particularly interesting. We attribute this to the fact that among other things, Flan-T5 was trained in arithmetic reasoning. QDCAT~\cite{chen-etal-2020-question} is the current state-of-the-art in the DROP task. It was built for reading comprehension and is based on RoBERTa~\cite{liu_roberta}, but adds an additional question-conditioned reasoning step on top (using a graph-attention network).

\paragraph{Inference-on-Tables.}
Table~\ref{tab:final_evaluation_infotabs} presents the prediction accuracies (EM score) achieved on the InfoTabs~\cite{gupta-etal-2020-infotabs} dataset.

\begin{table}[ht]
\centering
\resizebox*{\linewidth}{!}{  
\begin{tabular}{llrrr}
\multicolumn{1}{l}{} &  & \multicolumn{1}{c}{\textbf{In-Domain}} & \multicolumn{1}{c}{\textbf{Cross-Domain}} & \multicolumn{1}{c}{\textbf{Adversarial}} \\ \hline
\multirow{2}{*}{\textbf{BART}} & Baseline & 33.30 & 23.67 & 27.68 \\
 & Ours & \multicolumn{1}{r}{\textbf{67.20}} &   \textbf{54.40} & \textbf{57.20} \\ \hline
 \multirow{2}{*}{\textbf{T5}} & Baseline & 32.00 & 11.76 & 13.00 \\
 & Ours & \multicolumn{1}{r}{32.30} &   \textbf{18.07} & \textbf{15.25} \\ \hline
\multirow{2}{*}{\textbf{Flan-T5}} & Baseline & \multicolumn{1}{r}{27.23} & 25.14 & 29.17 \\
 & Ours & \textbf{34.04} & 26.14 & 29.04 \\ \hhline{=====}
 \textbf{BPR} & & 78.42 & 71.97 & 70.03
\end{tabular}
}
\caption{Evaluation on the InfoTabs dataset. Our approach significantly improves the results on the in-domain data.}
\label{tab:final_evaluation_infotabs}
\end{table}

Similarly to reading comprehension, Arithmetic-Based Pretraining significantly improves EM scores in all cases. This applies especially to the in-domain test set. For the other two test sets, our approach also shows improvements over the baselines (mostly for BART~\cite{lewis2019bart}), indicating to improve the model's robustness and capability to extrapolate to unseen data. We attribute performance differences to model sizes. Furthermore, analysis of the in-domain test results shows that T5 and Flan-T5 are biased toward predicting entailment. Since we observe this in both Baseline and Ours, we do not attribute this to how the data was preprocessed for the Inferable Number Prediction Task (Section~\ref{sec:inp_datasets}). This is different for BART. An analysis of the in-domain test results shows that the model correctly predicts $60.30\%$ of entailments, $75.50\%$ of contradictions, and $65.83\%$ of neutrals. BPR~\cite{neeraja-etal-2021-incorporating} is the current state-of-the-art in the InfoTabs task. It is based on BERT~\cite{devlin2018bert} but built for inference over tabular data. It provides an improved representation of the input data, is pretrained on MultiNLI~\cite{williams-etal-2018-broad}, and incorporates external knowledge. 

\paragraph{Table-to-Text Generation.} \label{sub_sec:human_evaluation}
For human evaluation\footnote{The human evaluation was conducted by one of the authors.}, we follow the approach used by \newcite{moosavi2021learning} for evaluating the results on SciGen. As this is very time-consuming, we only analyse $100$ random table-description pairs from each, the SciGen and WikiBio~\cite{Lebret_EMNLP2016} dataset, and also only from the BART~\cite{lewis2019bart} experiments. For SciGen, we use the results from the large split experiment\footnote{For SciGen, BART is the current state-of-the-art, and the baseline results of our human evaluation are comparable with those reported by \newcite{moosavi2021learning}. We are not aware of a comparable human evaluation for WikiBio. Appendix~\ref{sec:evaluation_automatic_metrics} shows a comparison of automatic metrics for both datasets.}.

For annotation, we break down each generated output to its corresponding statements (facts). We create one CSV file for each dataset that contains these statements in random order. This way, the annotator can not see whether a statement was generated by Ours (BART with Arithmetic-Based Pretraining) or Baseline (BART without Arithmetic-Based Pretraining). Alongside with the generated statements, this CSV file contains the original tables and gold descriptions. The annotator then decides for each of the statements whether it belongs to one of the following labels:

\begin{itemize}
  \item \textit{Entailed}: The statement is entailed in the gold description, e.g., a fact that is mentioned either in a similar or different wording in the description.
  \item \textit{Extra}: The statement is not entailed in the gold description but is factually correct based on the table's content.
  \item \textit{Incorrect}: The statement is relevant to the table, i.e., it contains relevant entities but is factually incorrect. For instance, the statement says \textit{system A outperforms system B by 2 points} while based on the table system A has a lower performance than system B.
  \item \textit{Hallucinated}: The statement is not relevant to the table.
\end{itemize}

Based on these labels, we then compute the recall ($\text{\#entailed} / \text{\#gold}$), \mbox{precision} ($\text{\#entailed} / \text{\#generated}$), correctness ($(\text{\#entailed}+\text{\#extra}) / \text{\#generated}$), and \mbox{hallucination} ($\text{\#hallucinated} / \text{\#generated}$) scores for the generated facts. \#gold and \#generated refers to the respective number of included statements, not complete sequences. Table~\ref{tab:human_evaluation} shows the results.

\begin{table}[ht]
  \centering

\resizebox*{0.7\linewidth}{!}{
\begin{tabular}{lrrrr}
\multicolumn{1}{c}{} & \multicolumn{1}{c}{\textbf{Prec.}} & \multicolumn{1}{c}{\textbf{Rec.}} & \multicolumn{1}{c}{\textbf{Cor.}} & \multicolumn{1}{c}{\textbf{Hall.}} \\ \hline
\multicolumn{5}{c}{\textbf{SciGen}} \\ \hline
Baseline & 0.08 &  0.02 & 0.31 & 0.29 \\ 
Ours & 0.09 & 0.03 & \textbf{0.40} & 0.33 \\ \hline 
\multicolumn{5}{c}{\textbf{WikiBio}} \\ \hline
Baseline & 0.22 & 0.07 & 0.33 & 0.03 \\ 
Ours & \textbf{0.28} & \textbf{0.09} & \textbf{0.46} & 0.02 \\ 

\end{tabular}}

\caption{Results of the human evaluation. In both cases, our approach improves the correctness of the generated facts.}
\label{tab:human_evaluation}
\end{table}

Arithmetic-Based Pretraining improves the precision, recall, and correctness for both SciGen and WikiBio. In case of WikiBio, it improves the precision by $0.06$ points, suggesting that generated statements are more concise and closer to the target description. It also improves the ratio of statements that are factually correct by $0.13$ points. 
In case of SciGen, the baseline results reflect the results reported by \newcite{moosavi2021learning}, who also used the large variant of BART for their experiments. Ours improves the results in almost every aspect (especially in case of factual correctness, where it improves the results by $0.09$ points). However, we observe a slight increase in hallucinations, which is a minor deterioration. We found that while Baseline seems to generate descriptions close to the target, Ours is somewhat more oriented towards the tabular values, whereby these values are used out-of-context in some cases which might be the reason for this deterioration. Nevertheless, all models generate fluent and valid-looking descriptions (see Appendix~\ref{sec:appendix_5} for examples). This suggests that Arithmetic-Based Pretraining has no negative impact on a model's text generation capability. This is also supported by the results achieved using automatic metrics (see Appendix~\ref{sec:evaluation_automatic_metrics}).

\subsection{Out-of-Domain Pretraining}
\label{sec:ood_data}
To investigate whether the effectiveness of Arithmetic-Based Pretraining is a result of using in-domain data for pretraining (domain-adaptive pretraining) or improved numeracy, we evaluate our approach using out-of-domain data for pretraining. 
We focus on BART~\cite{lewis2019bart} for this experiment and perform Arithmetic-Based Pretraining on a different dataset before finetuning on DROP~\cite{dua-etal-2019-drop} and InfoTabs~\cite{gupta-etal-2020-infotabs}. For instance, for the DROP experiments, we pretrain models on WikiBio~\cite{Lebret_EMNLP2016}, SciGen~\cite{moosavi2021learning}, and InfoTabs, which all include data from a different domain, before finetuning. For SciGen, we use the large split in this experiment. 

\begin{table}[ht]
  \centering
  \resizebox*{0.7\linewidth}{!}{
\begin{tabular}{lrr}
 & \multicolumn{1}{l}{\textbf{EM}} & \multicolumn{1}{l}{\textbf{F1}} \\ \hline
\multicolumn{3}{c}{\textbf{DROP}} \\ \hline
DROP (in-domain) & 45.60 & 49.50 \\
Wikibio $\rightarrow$ DROP & 6.00 & 33.50 \\
InfoTabs $\rightarrow$ DROP & 35.50 & 39.63 \\
SciGen $\rightarrow$ DROP & \textbf{47.70} & \textbf{51.60} \\ \hline

\multicolumn{3}{c}{\textbf{InfoTabs}} \\ \hline
InfoTabs (in-domain) & \textbf{67.20} & - \\
WikiBio $\rightarrow$ InfoTabs & 33.15 & - \\
DROP $\rightarrow$ InfoTabs & 32.80 & - \\
SciGen $\rightarrow$ InfoTabs & 64.70 & -
\end{tabular}}
  \caption{Results of the out-of-domain pretraining (see Tables~\ref{tab:final_evaluation_drop} and~\ref{tab:final_evaluation_infotabs} for the in-domain experiments).}
  \label{tab:cross_evaluation}
  \end{table}

Table~\ref{tab:cross_evaluation} shows the results. Overall, the models pretrained using SciGen achieve the best out-of-domain results in both cases. In case of DROP, the results even exceed the ones achieved with in-domain pretraining. We find that the extent to which the pretraining dataset requires understanding and working with numbers has a major impact on the downstream performance (the more, the greater the impact). Among the datasets used, SciGen is in particular designed for the task of text generation based on arithmetic reasoning. It has a high number-to-word ratio and the subset used for pretraining on the Inferable Number Prediction Task (see Section~\ref{sec:mnp}) predominantly depends on arithmetic operations such as multiplications or divisions (see Table~\ref{tab:arithmetic_operations}) instead of lookups or orderings (like in the other datasets).
\section{Ablation Study} \label{sec:intrinsic_evaluation}
In this section, we investigate the impact of Arithmetic-Based Pretraining on the numeracy of a pretrained language model. Due to the shortcomings of automatic metrics in table-to-text generation (see Section~\ref{sec:metrics}) and because we want to be able to compare and discuss the impact of each component across datasets, we use the Inferable Number Prediction task for this and evaluate the number of correctly predicted context-related masked numbers (please see Appendix~\ref{sec:ablation_downstream} for ablation experiments in downstream tasks)\footnote{In case of the contrastive loss, we also experiment with other number representations (see Appendix~\ref{sec:appendix_4}).}. We use the preprocessed subsets of the original datasets for the Inferable Number Prediction Task (see Section~\ref{sec:inp_datasets}). For evaluation, we use Exact Match (EM score) and F1 score (see Section~\ref{sec:metrics}). \mbox{Table}~\ref{tab:results_masked_word} shows the results. 

\begin{table}[ht]
  \centering
  \resizebox*{0.5\linewidth}{!}{
\begin{tabular}{lrr}
 & \textbf{EM} & \textbf{F1} \\ \hline
\multicolumn{3}{c}{\textbf{WikiBio}} \\ \hline
BART & 29.69 & 48.12 \\
CLT + INP & \textbf{43.13} & \textbf{69.97} \\
Ours & \textbf{77.38} & \textbf{74.69}\\
\hline
\multicolumn{3}{c}{\textbf{SciGen}} \\ \hline
BART & 7.04 & 32.21 \\
DT + INP & 7.20 & \textbf{35.11} \\
CLT + INP &  \textbf{12.26} & \textbf{36.78} \\
Ours & \textbf{24.68} & \textbf{45.81}\\
Ours - INP & \textbf{21.49} & \textbf{40.51} \\
\hline
\multicolumn{3}{c}{\textbf{InfoTabs}} \\ \hline
BART & 12.43 & 22.17 \\
DT + INP & \textbf{23.20} & \textbf{46.17} \\
CLT + INP & \textbf{59.09} & \textbf{73.88} \\
Ours & \textbf{60.45} & \textbf{74.33}\\
Ours - INP & \textbf{59.66} & \textbf{72.71} \\
\hline
\multicolumn{3}{c}{\textbf{DROP}} \\ \hline
BART & 7.20 & 7.20 \\
DT + INP & 6.33 & \textbf{55.51} \\
CLT + INP & \textbf{29.40} & \textbf{66.43} \\ 
Ours & \textbf{30.58} & \textbf{67.07}\\
Ours - INP & \textbf{25.37} & \textbf{59.83} \\
\end{tabular}}

\caption{Ablation study on the Inferable Number Prediction Task. We conduct DT + INP and Ours - INP once for each task and with SciGen~\cite{moosavi2021learning} as representative for table-to-text generation.}
\label{tab:results_masked_word}
\end{table}

We consider the large variant of BART~\cite{lewis2019bart} with its default tokenisation (DT) and masking procedure (DM) as baseline for this \mbox{experiment}. \emph{DT + INP} uses the default tokenisation but our masking procedure (INP). 
\emph{CLT + INP} then uses the character-level tokenisation for numbers (CLT). \emph{Ours} finally combines CLT and INP with the contrastive loss (CL) as supporting signal to improve the representation of numbers. As last ablation, \emph{Ours - INP} combines CLT with the contrastive loss but uses DM instead of INP and shows the contribution of our masking procedure to the effectiveness of Arithmetic-Based Pretraining. 

In comparison with BART, DT + INP shows that our masking procedure improves the results across all tasks. This is most significant in case of InfoTabs (up to $10.77$ points in EM score). In case of DROP, it raises the F1 score from $7.20$ to $55.51$ points, meaning that there is a significantly larger overlap between predicted numbers and target numbers. Using character-level instead of default tokenisation for numbers (CLT + INP) again improves the results across all datasets, indicating improved capabilities for arithmetic operations. Compared to DT + INP, it improves the EM score by $35.89$ points in case of InfoTabs, and by $23.07$ points in case of DROP. Ours further improves the results across all datasets. This is most significant in case of the table-to-text datasets, where it improves the EM score by 34.25 points in case of WikiBio~\cite{Lebret_EMNLP2016}, and 12.42 points in case of SciGen~\cite{moosavi2021learning}. Since we create the pairs for the contrastive loss batch-wise, i.e., we consider all numbers in a batch independently from the samples (see Section~\ref{sec:contrastive}), an advantageous number-to-word ratio favors a good positive-negative pair ratio for the contrastive loss, as in the case of SciGen which has the highest number to word ratio in input tables ($33.88\%$, see also Section~\ref{sect:dataset}). This is counteracted by WikiBio which has a lower number-to-word ratio ($16.32\%$). However, with $728,321$ samples, Wikibio is the largest dataset. We therefore assume that more data compensates for a poor number-to-word ratio. Ours - INP deteriorates the EM score by $5.21$ points in case of DROP, $3.19$ points in case of SciGen, and $0.79$ points in case of InfoTabs. This shows the contribution of our masking procedure to the effectiveness of Arithmetic-Based Pretraining. 
\section{Conclusions}
In this paper, we propose Arithmetic-Based Pretraining, an approach for jointly addressing the shortcomings of pretrained language models in understanding and working with numbers (usually referred to as numeracy). In contrast to existing approaches, Arithmetic-Based Pretraining does not require architectural changes or pretraining from scratch. It uses contrastive learning to improve number representation and a novel extended pretraining objective, the Inferable Number Prediction Task, to improve numeracy in just one extended pretraining step. Our experiments show performance improvements due to better numeracy in three different state-of-the-art pretrained language models, BART, T5, and Flan-T5, across various tasks and domains, including reading comprehension (DROP), inference-on-tables (InfoTabs), and table-to-text generation (SciGen and WikiBio). We show that the effectiveness of our approach is not limited to in-domain pretraining, but rather depends on the extent to which the dataset used in the Inferable Number Prediction Task requires understanding numbers. For example, pretraining on the SciGen dataset improves the results achieved on DROP. Our ablation studies show that contrastive learning and the Inferable Number Prediction Task are key to improving the numeracy of the examined models.
\section{Limitations}
Our work is subject to some limitations. First of all, due to hardware limitations, we could not use the large variant of T5~\cite{raffel_t5} and Flan-T5~\cite{chung_flan_t5} in a setting comparable to our BART-large experiments. Furthermore, BART~\cite{lewis2019bart} restricts the maximum length of input sequences to $1024$ characters\footnote{\url{https://huggingface.co/docs/transformers/model_doc/bart\#transformers.BartConfig}, last accessed on 10/02/23.}. For better comparability, we also use T5 and Flan-T5 accordingly. This limitation is due to the increased computational complexity of longer input sequences, but it is problematic with table-to-text generation datasets. For example, SciGen~\cite{moosavi2021learning} consists in large parts of tables that exceed this sequence length when represented as a linearized sequence. While we have tried to take this into account by reducing the input data to necessary information, it was not guaranteed that the model always sees the complete information, which certainly has a negative impact on the evaluation results achieved on the downstream tasks. We guess that the results would have been more expressive if we would have used a different representation for tables, or focused on models that do not have this sequence length limitation. 

Another limitation of our work concerns the impact of contrastive learning. According to \newcite{henderson2017efficient}, the impact of contrastive loss is favored by large batch sizes. Due to hardware limitations, we were only able to use small batch sizes (see Appendix~\ref{sec:appendix_3}). The models might have adapted better if we would had the possibility to train with larger batch sizes. Regarding the weighting of contrastive and masked loss in the joint loss function, we only use equal weighting for our experiments, since we found that this already leads to good results, and due to the already large number of experiments conducted in this paper, we did not experiment with other weightings. However, optimizing this hyperparameter could further improve the results.

Evaluation is also a critical point. Although metrics such as PARENT~\cite{dhingra2019handling} try to measure the factual correctness of generated descriptions, it requires a more individual examination in many cases. Especially in such highly specialized scenarios such as SciGen. Therefore, we conduct a human evaluation in order to analyse the impact of our Arithmetic-Based Pretraining on the downstream tasks. However, due to limited resources, we were only able to conduct a small-scale human evaluation. At this point, we would also like to mention that our evaluation setup in general is subject to limitations. As an extended pretraining approach, Arithmetic-Based Pretraining might have a negative impact on a model's general applicability, i.e., downstream performance in tasks used for pretraining, e.g., translation in case of T5, or other non-number related tasks commonly used in model benchmarking, such as question answering, text classification, or sentiment analysis. We only examined the impact on text generation as part of our human evaluation and with automatic metrics (see Appendix~\ref{sec:evaluation_automatic_metrics}). However, since (1) the Inferable Number Prediction Task (Section~\ref{sec:mnp}) is a variation of the widely used masked language modeling objective~\cite{devlin2018bert}, and (2) character-level tokenisation does not introduce new embeddings into a pretrained language model, we don't expect a negative impact here. 

Another limitation concerns the evaluation of the Inferable Number Prediction Task on a model's numeracy. Since it is not reliably traceable whether and which arithmetic operation was used by a model to come to a specific result, we can only infer improved capabilities for arithmetic operations by performance improvements in the Inferable Number Prediction Task. We cannot clearly distinguish performance improvements on specific arithmetic operations. 

\section{Acknowledgements}

This research work has been funded by the German Federal Ministry of Education and Research and the Hessian Ministry of Higher Education, Research, Science and the Arts within their joint support of the National Research Center for Applied Cybersecurity ATHENE. It also received funding from the German Research Foundation (DFG) under grant № EC 503/1-1 and GU 798/21-1.

\bibliography{anthology,custom}
\bibliographystyle{acl_natbib}

\appendix

\section{Hyperparameters for Experiments}
\label{sec:appendix_3}

Table~\ref{tab:experiments_setup} shows the hyperparameter configuration for our experiments. In order to not train longer than necessary, we have determined the optimal number of epochs for each experiment by using early stopping with a patience of $10$. For the downstream tasks, we have used the MoverScore~\cite{zhao2019moverscore} with the table-to-text generation datasets. For DROP~\cite{dua-etal-2019-drop} and InfoTabs~\cite{gupta-etal-2020-infotabs}, we have used the EM score. All models were trained for the same amount of epochs.

\begin{table}[H]  
  \centering
  
\resizebox*{0.8\linewidth}{!}{  
\begin{tabular}{lrrr}

\multicolumn{1}{l}{\textbf{}} & \multicolumn{1}{l}{\textbf{Batch Size}} & \multicolumn{1}{l}{\textbf{Epochs}} & \multicolumn{1}{l}{\textbf{Learning Rate}} \\ \hline
\multicolumn{4}{c}{\textbf{Inferable Number Prediction Task}} \\ \hline
\multicolumn{1}{l}{SciGen} & \multicolumn{1}{r}{8} & \multicolumn{1}{r}{50} & 3e-5 \\ 
\multicolumn{1}{l}{WikiBio} & \multicolumn{1}{r}{8} & \multicolumn{1}{r}{3} & 3e-5 \\ 
\multicolumn{1}{l}{InfoTabs} & \multicolumn{1}{r}{8} & \multicolumn{1}{r}{21} & 3e-5 \\ 
\multicolumn{1}{l}{DROP} & \multicolumn{1}{r}{8} & \multicolumn{1}{r}{48} & 3e-5 \\ \hline
\multicolumn{4}{c}{\textbf{Downstream Tasks}} \\ \hline
\multicolumn{1}{l}{SciGen} & \multicolumn{1}{r}{8} & \multicolumn{1}{r}{27} & 3e-5 \\
\multicolumn{1}{l}{WikiBio} & \multicolumn{1}{r}{8} & \multicolumn{1}{r}{9} & 3e-5 \\
\multicolumn{1}{l}{InfoTabs} & \multicolumn{1}{r}{8} & \multicolumn{1}{r}{14} & 3e-5 \\
\multicolumn{1}{l}{DROP} & \multicolumn{1}{r}{8} & \multicolumn{1}{r}{10} & 3e-5 \\
\end{tabular}
}

  \caption{Hyperparameter Configuration.}
  \label{tab:experiments_setup}
\end{table}
\section{Inferable Number Prediction Task -- Example Input Data}
\label{sec:masked_task_examples}
\begin{figure}[H]
  \centering
  \includegraphics[width=0.9\linewidth]{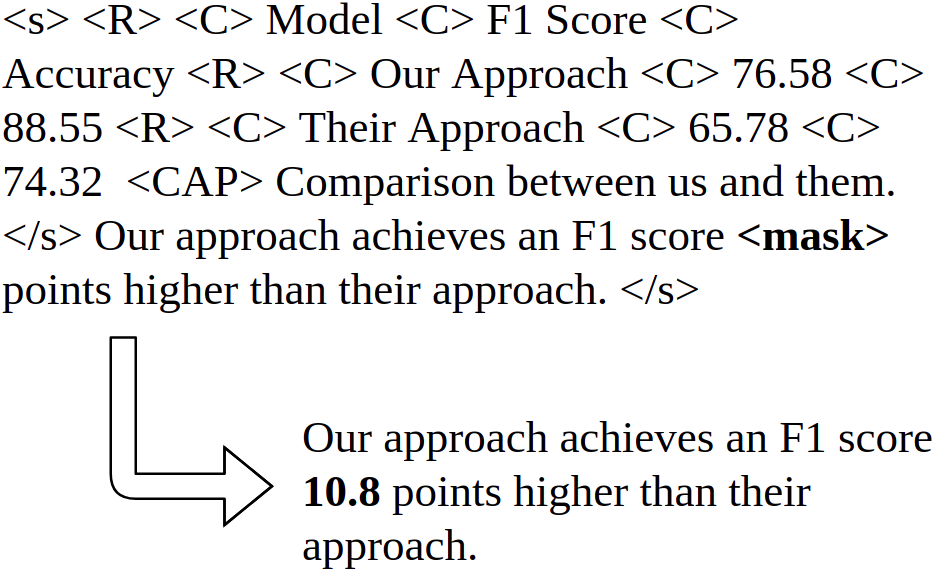}
  \caption{Illustration of a linearized table that is used for the Inferable Number Prediction Task. \textit{<R>}, \textit{<C>} and \textit{<CAP>} symbolize the beginning of a new row, cell, and the table's caption.}
  \label{fig:masked_example}
  \end{figure}

For table-to-text generation, Figure~\ref{fig:masked_example} shows an example of a (linearized) table from SciGen~\cite{moosavi2021learning} with its caption as $C_1$, concatenated to its masked description $C_2$ using \textit{</s>}. \textit{<s>} and \textit{</s>} are special tokens used by BART~\cite{lewis2019bart} to represent the beginning and ending of a sequence. In case of WikiBio~\cite{Lebret_EMNLP2016}, the input data is represented accordingly.

\begin{figure}[H]
  \centering
  \includegraphics[width=0.9\linewidth]{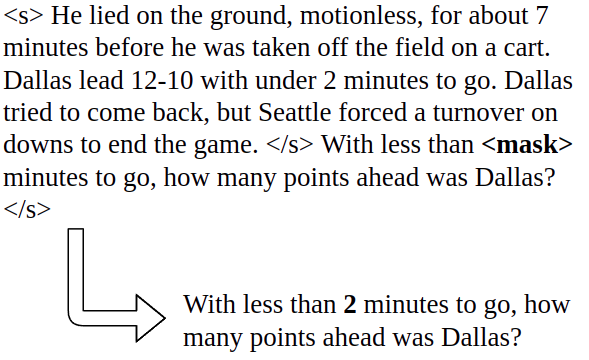}
  \caption{Illustration of an input sample for the Inferable Number Prediction Task using DROP.}
  \label{fig:drop_example}
  \end{figure}

For DROP~\cite{dua-etal-2019-drop}, Figure~\ref{fig:drop_example} shows an example. It consists of the paragraph $C_1$, and a question $C_2$. The question contains a number ($2$) that also occurs in the paragraph. 

\begin{figure}[H]
  \centering
  \includegraphics[width=0.9\linewidth]{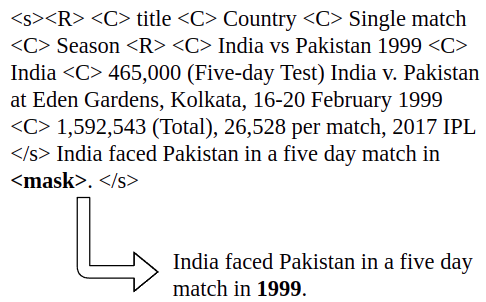}
  \caption{Illustration of an input sample for the Inferable Number Prediction Task using InfoTabs.}
  \label{fig:infotabs_example}
  \end{figure}

Figure~\ref{fig:infotabs_example} shows an example for the InfoTabs~\cite{gupta-etal-2020-infotabs} datasets. It is basically the same as for the table-to-text generation datasets, but uses the hypothesis as $C_2$.

\section{Inferable Number Prediction Task -- Dataset Details}
\label{sec:masked_task_arithmetic_details}
In this section, we want to provide more details on the distribution of arithmetic operations across datasets used for the Inferable Number Prediction Task. Table~\ref{tab:infotabs_details} shows the ratio of each arithmetic operation on the overall number of samples for each split for the InfoTabs~\cite{gupta-etal-2020-infotabs} dataset.

\begin{table}[htb]
\resizebox*{1.0\linewidth}{!}{  
\begin{tabular}{lrrrrrr}
 & \multicolumn{1}{c}{\textbf{OCC}} & \multicolumn{1}{c}{\textbf{ORD}} & \multicolumn{1}{c}{\textbf{SUM}} & \multicolumn{1}{c}{\textbf{SUB}} & \multicolumn{1}{c}{\textbf{MUL}} & \multicolumn{1}{c}{\textbf{DIV}} \\ \hline
\textbf{Train} & 0.24 & 0.35 & 0.05 & 0.16 & 0.15 & 0.05 \\
\textbf{Dev} & 0.15 & 0.34 & 0.07 & 0.18 & 0.20 & 0.06 \\
\textbf{Test} & 0.22 & 0.16 & 0.09 & 0.23 & 0.23 & 0.07
\end{tabular}}
\caption{Ratio of arithmetic operations for each split of the InfoTabs dataset.}
\label{tab:infotabs_details}
\end{table}

Table~\ref{tab:drop_details} shows this ratio for the DROP~\cite{dua-etal-2019-drop} dataset.

\begin{table}[htb]
\resizebox*{1.0\linewidth}{!}{  
\begin{tabular}{lrrrrrr}
 & \multicolumn{1}{c}{\textbf{OCC}} & \multicolumn{1}{c}{\textbf{ORD}} & \multicolumn{1}{c}{\textbf{SUM}} & \multicolumn{1}{c}{\textbf{SUB}} & \multicolumn{1}{c}{\textbf{MUL}} & \multicolumn{1}{c}{\textbf{DIV}} \\ \hline
\textbf{Train} & 0.41 & 0.32 & 0.4 & 0.07 & 0.13 & 0.03 \\
\textbf{Dev} & 0.42 & 0.31 & 0.05 & 0.05 & 0.14 & 0.03 \\
\textbf{Test} & 0.43 & 0.30 & 0.04 & 0.05 & 0.15 & 0.03
\end{tabular}}
\caption{Ratio of arithmetic operations for each split of the DROP dataset.}
\label{tab:drop_details}
\end{table}

Table~\ref{tab:scigen_details} shows this ratio for the SciGen~\cite{moosavi2021learning} dataset.

\begin{table}[htb]
\resizebox*{1.0\linewidth}{!}{ 
\begin{tabular}{lrrrrrr}
 & \multicolumn{1}{c}{\textbf{OCC}} & \multicolumn{1}{c}{\textbf{ORD}} & \multicolumn{1}{c}{\textbf{SUM}} & \multicolumn{1}{c}{\textbf{SUB}} & \multicolumn{1}{c}{\textbf{MUL}} & \multicolumn{1}{c}{\textbf{DIV}} \\ \hline
\textbf{Train} & 0.11 & 0.06 & 0.04 & 0.12 & 0.40 & 0.27 \\
\textbf{Dev} & 0.11 & 0.05 & 0.04 & 0.12 & 0.43 & 0.25 \\
\textbf{Test} & 0.15 & 0.09 & 0.02 & 0.19 & 0.43 & 0.13
\end{tabular}}
\caption{Ratio of arithmetic operations for each split of the SciGen dataset.}
\label{tab:scigen_details}
\end{table}

Table~\ref{tab:wikibio_details} shows this ratio for the WikiBio~\cite{Lebret_EMNLP2016} dataset.

\begin{table}[!htb]
\resizebox*{1.0\linewidth}{!}{
\begin{tabular}{lrrrrrr}
 & \multicolumn{1}{c}{\textbf{OCC}} & \multicolumn{1}{c}{\textbf{ORD}} & \multicolumn{1}{c}{\textbf{SUM}} & \multicolumn{1}{c}{\textbf{SUB}} & \multicolumn{1}{c}{\textbf{MUL}} & \multicolumn{1}{c}{\textbf{DIV}} \\ \hline
\textbf{Train} & 0.25 & 0.38 & 0.03 & 0.10 & 0.20 & 0.03 \\
\textbf{Dev} & 0.25 & 0.38 & 0.03 & 0.10 & 0.19 & 0.04 \\
\textbf{Test} & 0.25 & 0.38 & 0.03 & 0.11 & 0.20 & 0.03
\end{tabular}}
\caption{Ratio of arithmetic operations for each split of the SciGen dataset.}
\label{tab:wikibio_details}
\end{table}
\section{Evaluation Using Automatic Metrics}
\label{sec:evaluation_automatic_metrics}

This section presents the evaluation of our results on table-to-text datasets using automatic metrics. For this, we use a variety of metrics commonly used for this task, i.e., \textit{BLEU}~\cite{papineni2002bleu}, \textit{MoverScore}~\cite{zhao2019moverscore}, \textit{BLEURT}~\cite{sellam2020bleurt}, and \textit{PARENT}~\cite{dhingra2019handling}. While BLEU calculates the concordance between the predicted description and the actual target on word-level, MoverScore and BLEURT measure the semantic concordance between the predicted description and the target using BERT~\cite{devlin2018bert}. BLEURT also takes the fluency of the predictions into account. PARENT estimates the factual correctness by comparing the predicted description to the original table and the target description, and especially rewards correct information that is contained in the table but not in the target. It has a higher correlation with human judgment. Table~\ref{tab:final_evaluation} reports the results. We highlight statistically significant improvements of our approach over the respective baseline in the tables (independent two-sample t-test, $p \leq 0.05$).

\begin{table}[ht]
  \centering
  \resizebox*{\linewidth}{!}{\begin{tabular}{lllrrrr}
    
    \multicolumn{3}{c}{} & \textbf{MoverS} & \textbf{BLEU} & \textbf{BLEURT} & \textbf{PARENT} \\ \hline
    \multicolumn{7}{c}{\textbf{SciGen}} \\ \hline
    \multicolumn{1}{l}{\multirow{6}{*}{{BART}}} 
    & \multicolumn{1}{l}{\multirow{3}{*}{Baseline}} 
    & {Few} & 52.48 & 4.60 & -0.63 & 3.38\\
    & & Medium & 53.76 & 4.26 & -0.69 & 3.72\\
    & & Large & 53.43 & 4.87 & -0.70 & 3.68\\ \cline{2-7}
    & \multicolumn{1}{l}{\multirow{3}{*}{\textbf{Ours}}} 
    & {Few} & \textbf{53.30} & 1.73 & -0.76 & 3.81 \\
    & & Medium & \textbf{53.40} & 2.71 & -0.78 & 3.45\\
    & & Large & \textbf{55.00} & \textbf{9.30} & -0.76 & 3.82\\ \hhline{~======}
    & BART (Moosavi et al.) & Large & 14.00 & 5.04 & -0.71 & - \\ 
    \hline
    \multicolumn{1}{l}{\multirow{6}{*}{{T5}}} 
    & \multicolumn{1}{l}{\multirow{3}{*}{Baseline}} 
    & {Few} & 52.30 & 2.96 & -0.94 & 6.39 \\
    & & Medium & 51.79 & 2.67 & -0.95 & 4.08 \\
    & & Large & 53.00 & 3.40 & -0.70 & 5.18\\ \cline{2-7}
    & \multicolumn{1}{l}{\multirow{3}{*}{\textbf{Ours}}} 
    & {Few} & 52.00 & 2.83 & -0.98 &  4.32\\
    & & Medium & 52.00 & 2.51 & -0.86 &  4.70\\
    & & Large & 53.40 & 2.96 & -0.89 &  \textbf{6.72}\\ \hhline{~======}
    & BART (Moosavi et el.) & Large & 6.00 & 3.38 & -0.79 & - \\ \hline 
\multicolumn{1}{l}{\multirow{6}{*}{{Flan-T5}}} 
    & \multicolumn{1}{l}{\multirow{3}{*}{Baseline}} 
    & {Few} & 53.03 & 2.76 & -0.67 & 7.89 \\
    & & Medium & 53.56 & 3.03 & -0.68 & 6.14 \\
    & & Large & 54.15 & 3.54 & -0.65 & 7.94 \\ \cline{2-7}
    & \multicolumn{1}{l}{\multirow{3}{*}{\textbf{Ours}}} 
    & {Few} & \textbf{54.22} & 3.14 & -0.65 &  8.54 \\
    & & Medium & \textbf{54.76} & 3.25 & -0.71 &  \textbf{8.12} \\
    & & Large & \textbf{55.12} & 3.34 & -0.61 &  \textbf{9.32}\\  \hline     

    \multicolumn{7}{c}{\textbf{WikiBio}} \\ \hline
    \multicolumn{1}{l}{\multirow{2}{*}{{BART}}} & 
    Baseline & & 61.50 & 17.98 & -0.64 & 45.18 \\
    & \textbf{Ours} & & \textbf{62.78} & 18.54 & -0.27 & 44.32 \\\cline{2-7}
    \multicolumn{1}{l}{\multirow{2}{*}{{T5}}} & 
    Baseline & & 60.30 & 17.94 & -0.86 & 43.97 \\
    & \textbf{Ours} & & 60.10 & 20.00 & -0.22 &  \textbf{45.25}\\ \cline{2-7}
    \multicolumn{1}{l}{\multirow{2}{*}{{Flan-T5}}} & 
    Baseline & & 59.81 & 17.56 & -0.78 & 44.67 \\
    & \textbf{Ours} & & \textbf{62.51} & \textbf{21.11} & \textbf{-0.18} &  \textbf{46.10}\\ \hhline{~======}
    & MBD & & - & 41.56 & - & 56.16 \\ 

    \end{tabular}}

  \caption{Evaluation of our results on table-to-text datasets using automatic metrics. \textit{Baseline} presents the results of the BART-large and Flan-T5-base models without Arithmetic-Based Pretraining. \textit{Ours} shows the results of these models with Arithmetic-Based Pretraining.}
  \label{tab:final_evaluation}
  \end{table}

The results show that Arithmetic-Based Pretraining slightly improves the performance in most experiments (based on PARENT and MoverScore), and has no negative impact text generation capabilities. However, as outlined in Section~\ref{sec:metrics},  none of these metrics can really assess the correctness of a fact that might be reasoned from the source data~\cite{moosavi2021learning, chen2020logical, suadaatowards}. PARENT tries to address this, which is why this metric is the most appropriate one. Like BLEURT, Moverscore measures the semantic concordance between target and prediction. The advantage of MoverScore is that it is easier to interpret.

In case of SciGen, even our baseline results for BART~\cite{lewis2019bart} are better than reported by \newcite{moosavi2021learning}. We attribute this to different training hyperparameters (they did not report hyperparameters). While BART~\cite{lewis2019bart} and T5~\cite{raffel_t5} are state-of-the-art in SciGen~\cite{moosavi2021learning}, MBD~\cite{mbd} is the state-of-the-art in WikiBio~\cite{Lebret_EMNLP2016}. It is a multi-branch decoder that was build to reduce the hallucination in data-to-text tasks.
\section{Ablation Study -- Downstream Tasks}
\label{sec:ablation_downstream}

This section shows the results of our downstream ablation experiments. For experiments, we use the same setup as described in Section~\ref{sec:intrinsic_evaluation}, i.e., we consider the large variant of BART~\cite{lewis2019bart} with its default tokenisation (DT) and masking procedure (DM) as baseline for this experiment. Additionally, we finetune the models in the downstream task (using the hyperparameters described in Appendix~\ref{sec:appendix_3}). For evaluation, we use the respective test splits (in-domain in case of InfoTabs~\cite{gupta-etal-2020-infotabs}). Table~\ref{tab:downstream_ablation_table_2_tex} and Table~\ref{tab:downstream_ablation_drop_infotabs} show the results of our ablation experiments in downstream tasks. We conduct the same experiments as for the general ablation study (Section~\ref{sec:intrinsic_evaluation}): \emph{DT + INP} uses the default tokenisation but our masking procedure (the Inferable Number Prediction Task, Section~\ref{sec:mnp}), \emph{CLT + INP} uses the character-level tokenisation for numbers (CLT), \emph{Ours} combines CLT and INP with the contrastive loss (CL), and \emph{Ours - INP} combines CLT with the contrastive loss but uses DM instead of INP. Overall, the results reflect the findings described in Section~\ref{sec:intrinsic_evaluation}. We highlight statistically significant improvements of our approach over the respective baseline in the tables (independent two-sample t-test, $p \leq 0.05$).

\begin{table}[ht]
  \centering
  \resizebox*{0.6\linewidth}{!}{
\begin{tabular}{lrr}
 & \textbf{MoverScore} & \textbf{BLEU} \\ \hline
\multicolumn{3}{c}{\textbf{WikiBio}} \\ \hline
BART & 61.50 & 17.98 \\
DT + INP & 61.74 & 17.31 \\
CLT + INP & \textbf{62.01} & \textbf{18.42} \\
Ours & \textbf{62.78} & \textbf{18.54} \\
Ours - INP & \textbf{62.15} & \textbf{18.25} \\
\hline
\multicolumn{3}{c}{\textbf{SciGen}} \\ \hline
BART & 53.43 & 4.87 \\
DT + INP & 53.76 & 4.65 \\
CLT + INP &  \textbf{54.12} & \textbf{6.45} \\
Ours & \textbf{55.00} & \textbf{9.30} \\
Ours - INP & \textbf{54.87} & \textbf{7.32} \\
\end{tabular}}

\caption{Downstream ablation study for SciGen and WikiBio.}
\label{tab:downstream_ablation_table_2_tex}
\end{table}

According to automatic metrics, the impact on table-to-text generation is rather limited. We suspect that this is partly due to their shortcomings in assessing the correctness of information not directly included in the source data (see also Section~\ref{sec:metrics}). DT + INP shows that pretraining using our masking procedure slightly improves the results in both cases. Using the character-level tokenisation for numbers further improves the results (CLT + INP). In case of SciGen, the comparison between Ours and Ours - INP suggests that using the character-level tokenisation and contrastive learning to improve the number representation has more impact than pretraining using INP. In case of WikiBio, the differences are rather negligible (although Ours outperforms the baseline). This might be due to the characteristics of the dataset. As described in Section~\ref{sect:dataset}, WikiBio rather requires copying numbers from input tables to output text, than inferring context-related numbers (which is different in the other datasets). 

\begin{table}[ht]
  \centering
  \resizebox*{0.5\linewidth}{!}{
\begin{tabular}{lrr}
 & \textbf{EM} & \textbf{F1} \\ \hline
\multicolumn{3}{c}{\textbf{DROP}} \\ \hline
BART & 36.00 & 39.26 \\
DT + INP & \textbf{39.87} & \textbf{43.77} \\
CLT + INP & \textbf{42.19} & \textbf{46.09} \\
Ours & \textbf{45.60} & \textbf{49.50} \\
Ours - INP & \textbf{43.68} & \textbf{47.45} \\
\hline
\multicolumn{3}{c}{\textbf{InfoTabs}} \\ \hline
BART & 33.30 & - \\
DT + INP & \textbf{48.21} & - \\
CLT + INP &  \textbf{61.56} & - \\
Ours & \textbf{67.20} & - \\
Ours - INP & \textbf{62.56} & - \\
\end{tabular}}

\caption{Downstream ablation study for DROP and InfoTabs}
\label{tab:downstream_ablation_drop_infotabs}
\end{table}

In case of DROP~\cite{dua-etal-2019-drop} and InfoTabs~\cite{gupta-etal-2020-infotabs}, the results are more expressive. In both cases, we find that just using INP (DT + INP) as an extended pretraining task already brings a significant improvement over the baselines. This is further improved by using character-level tokenisation for numbers (CLT + INP) and contrastive learning (Ours). Ours - INP shows that in both cases, INP has a significant impact on performance improvements.
\section{Experiments using other Contrastive Representations}
\label{sec:appendix_4}

Regarding the contrastive representation, we also experiment with number representations other than the default subword-level one in order to improve the representation of numbers using the character-level tokenisation, i.e., exponent-mantissa~\cite{zhang2020language}, a verbalized representation, and a combination of all of them using the Inferable Number Prediction Task. We focus on BART~\cite{lewis2019bart} (the large variant) for this experiment. We conduct this experiment using the large split of the SciGen dataset~\cite{moosavi2021learning}. Table~\ref{tab:contrastive_experiments} shows the results.  

\begin{table}[!htb]
  \centering
  \resizebox*{0.8\linewidth}{!}{
    \begin{tabular}{lrr}
        \multicolumn{1}{c}{\textbf{Experiment}} & \multicolumn{1}{c}{\textbf{EM}} & \multicolumn{1}{c}{\textbf{F1}} \\ \hline
        \begin{tabular}[c]{@{}l@{}}BART (verb. repr.) \end{tabular} & 15.69 & 41.01 \\ 
        \begin{tabular}[c]{@{}l@{}}BART (exp.-mant. repr)\end{tabular} & 18.13 & 36.78 \\ 
        \begin{tabular}[c]{@{}l@{}}BART (subword-based tok.)\end{tabular} & \textbf{24.68} & \textbf{45.81} \\ 
        \begin{tabular}[c]{@{}l@{}}BART (combined)\end{tabular} & 17.92 & 38.43 \\ 
        \end{tabular}
}

\caption{Comparison of results when using different representations for incorporating the character-level tokenisation.}
\label{tab:contrastive_experiments}
\end{table}

None of the other representations improves the results over using the default subword-level tokenisation. 
\section{Preliminary Math Experiments}
\label{sec:preliminary_math_experiments}

With GenBERT, \newcite{geva2020injecting} propose to start pretraining with math word problems in order to improve the model's number understanding and capabilities for arithmetic operations. Therefore, following this idea would be an obvious step in order to improve the numeracy of general purpose pretrained language models. Table~\ref{tab:mwp_results} shows the results of a preliminary experiment using GenBERT's math word problems dataset (MWP), BART~\cite{lewis2019bart}, and SciGen~\cite{moosavi2021learning} on the Inferable Number Prediction Task. We highlight statistically significant improvements of our approach over the respective baseline in the tables (independent two-sample t-test, $p \leq 0.05$).

\begin{table}[ht]
\centering
\resizebox*{\linewidth}{!}{  
\begin{tabular}{lrr}
 \multicolumn{1}{c}{\textbf{Experiment}}& \multicolumn{1}{c}{\textbf{EM}} & \multicolumn{1}{c}{\textbf{F1}} \\ \hline
Baseline & 7.20 & 35.11 \\
MWP-pretrained Baseline & \textbf{15.19} & 34.18 \\
MWP-pretrained Baseline + CLT & \textbf{22.94} & \textbf{42.55} \\
MWP-pretrained Baseline + CLT + CL & \textbf{22.78} & \textbf{43.14} \\
Ours & \textbf{24.68} & \textbf{45.81} \\ \hline
\end{tabular}
}
\caption{Results achieved on the Inferable Number Prediction Task with and without pretraining using math word problems.}
\label{tab:mwp_results}
\end{table}

\textit{Baseline} refers to the BART-large model. \textit{MWP-pretrained Baseline} shows the results for Baseline, but further pretrained on MWP. \textit{MWP-pretrained Baseline + CLT} represents the results for the MWP-pretrained Baseline, but uses the character-level representation (CLT) for numbers instead of BART's default tokenisation. Accordingly, \textit{MWP-pretrained Baseline + CLT + CL} incorporates the contrastive loss (CL) as additional training signal. The results show that pretraining using math word problems as a first step, in general, improves the results for the Inferable Number Prediction Task, but not over using Arithmetic-Based Pretraining (\textit{Ours}).

In case of SciGen, the Inferable Number Prediction Task, only uses samples with target descriptions that contain numbers that are inferable from the input table by lookup or arithmetic operations (see Section~\ref{sec:inp_datasets}). Therefore, even though it is a synthetic task, the results give insights on how effective pretraining on math word problems is for improving a model's numeracy.

\section{Examples from the Human Evaluation}
\label{sec:appendix_5}

\begin{figure*}[t]
  \centering
  \includegraphics[width=0.9\textwidth]{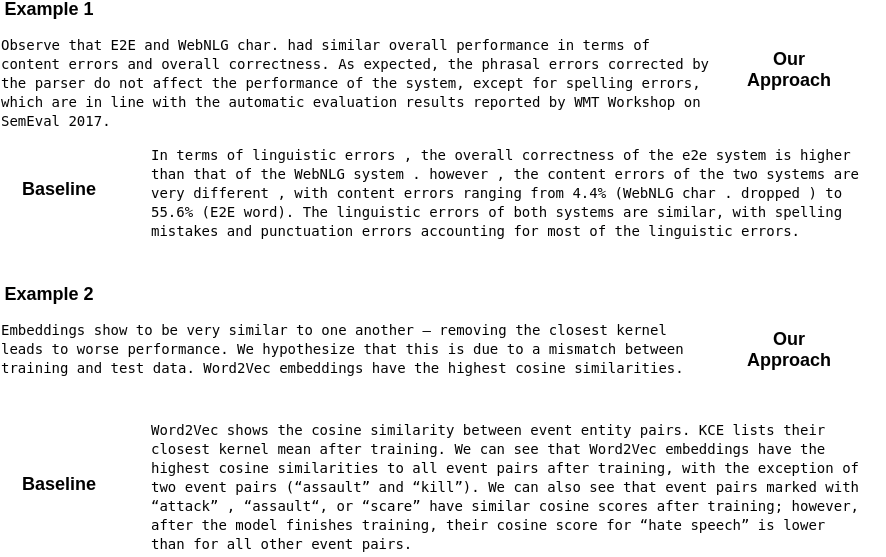}
  \caption{Generation from our approach and the BART baseline from the SciGen experiment using the medium split.}
  \label{fig:sample}
  \end{figure*}

Figure~\ref{fig:sample} shows two sample generations from our approach and the BART~\cite{lewis2019bart} baseline from the SciGen~\cite{moosavi2021learning} experiment using the medium split. Both read fluent and plausible.

\end{document}